\documentclass[10pt,twocolumn]{article}

\usepackage[margin=0.72in]{geometry}
\usepackage{cite}
\usepackage{amsmath,amssymb,amsfonts}
\usepackage{graphicx}
\usepackage{xcolor}
\usepackage{textcomp}
\usepackage{hyperref}
\usepackage{url}
\usepackage{booktabs}
\usepackage{multirow}
\usepackage{subcaption}
\usepackage{caption}
\usepackage{balance}
\usepackage{tikz}
\usetikzlibrary{calc}
\usetikzlibrary{positioning,fit,arrows.meta}
\hypersetup{
    colorlinks=true,
    linkcolor=blue,
    citecolor=blue,
    urlcolor=blue,
    pdftitle={DRIVE-C: A Controlled Corruption Dataset for Autonomous Driving},
    pdfauthor={Shiva Aher}
}

\title{DRIVE-C: A Controlled Corruption Dataset for Autonomous Driving}

\author{Shiva Aher\\
Georgia Institute of Technology, Georgia, United States\\
\texttt{saher8@gatech.edu}}

\date{arXiv preprint, May 2026. Under review at IEEE Data Descriptions.}

\begin{document}
\maketitle
\thispagestyle{empty}

\begin{abstract}
DRIVE-C is a controlled corruption dataset designed to evaluate visual perception robustness in autonomous driving systems. It is built from real-world forward-facing driving videos collected across daytime, nighttime, urban, rural, freeway, and parking environments. Clean clips are anonymized via localized face and license plate blurring, then transformed with physics-inspired synthetic degradations.

The dataset contains 10 clean clips and 600 corrupted clips spanning 12 camera degradation types across five severity levels, with per-clip metadata and Global Sensor Health Index (GSHI) annotations. DRIVE-C supports robustness benchmarking, degradation-aware modeling, uncertainty estimation, out-of-distribution (OOD) detection, and sensor health monitoring for Advanced Driver Assistance Systems (ADAS). By providing pixel-aligned clean and degraded video clips with fully reproducible corruption parameters, DRIVE-C offers a structured testbed for studying perception reliability under controlled camera degradation.
\end{abstract}

\noindent\textbf{Dataset DOI/PID:} 10.5281/zenodo.19656444\\
\textbf{Data type/location:} Video dataset with clean and synthetically corrupted clips, associated metadata, and GSHI annotations; collected in Michigan, United States; hosted on Zenodo.
\vspace{0.8em}

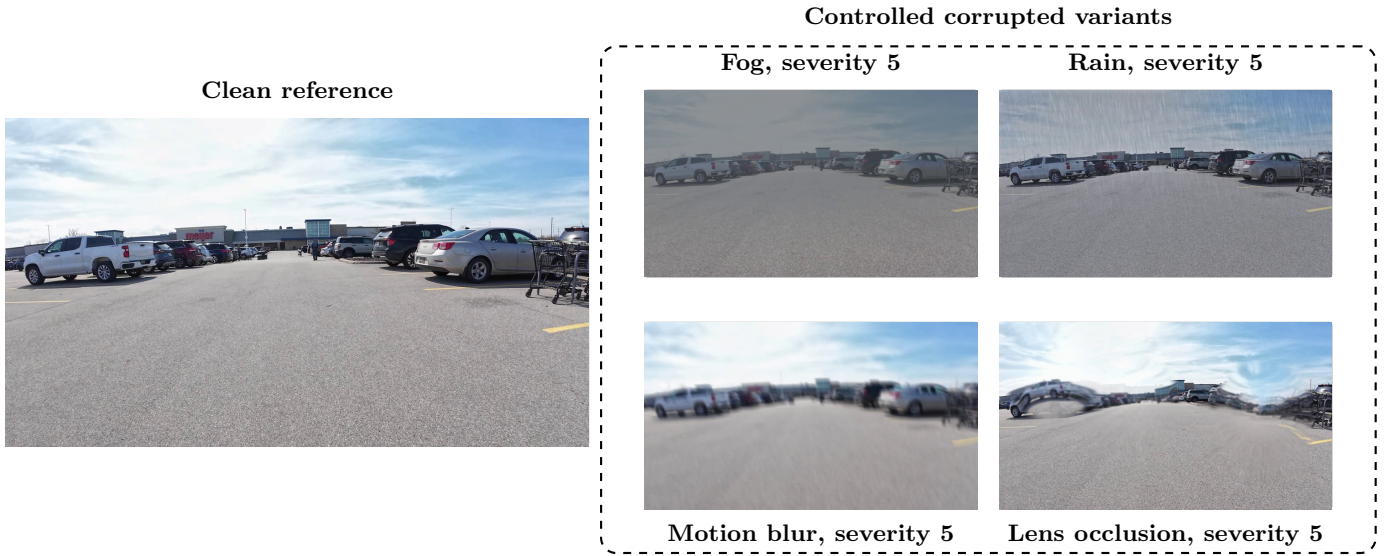
\begin{figure*}[!t]
\centering

\begin{tikzpicture}[
    font=\small,
    img/.style={
        draw=black!45,
        rounded corners=1pt,
        line width=0.4pt,
        inner sep=0pt,
        outer sep=0pt
    },
    label/.style={
        align=center,
        font=\small\bfseries
    },
    stat/.style={
        draw,
        rounded corners=1.5mm,
        thick,
        align=center,
        minimum height=8mm,
        minimum width=25mm,
        fill=blue!5,
        font=\scriptsize
    }
]

\node[img, anchor=center] (clean) at (0,0)
{\includegraphics[width=0.43\textwidth]{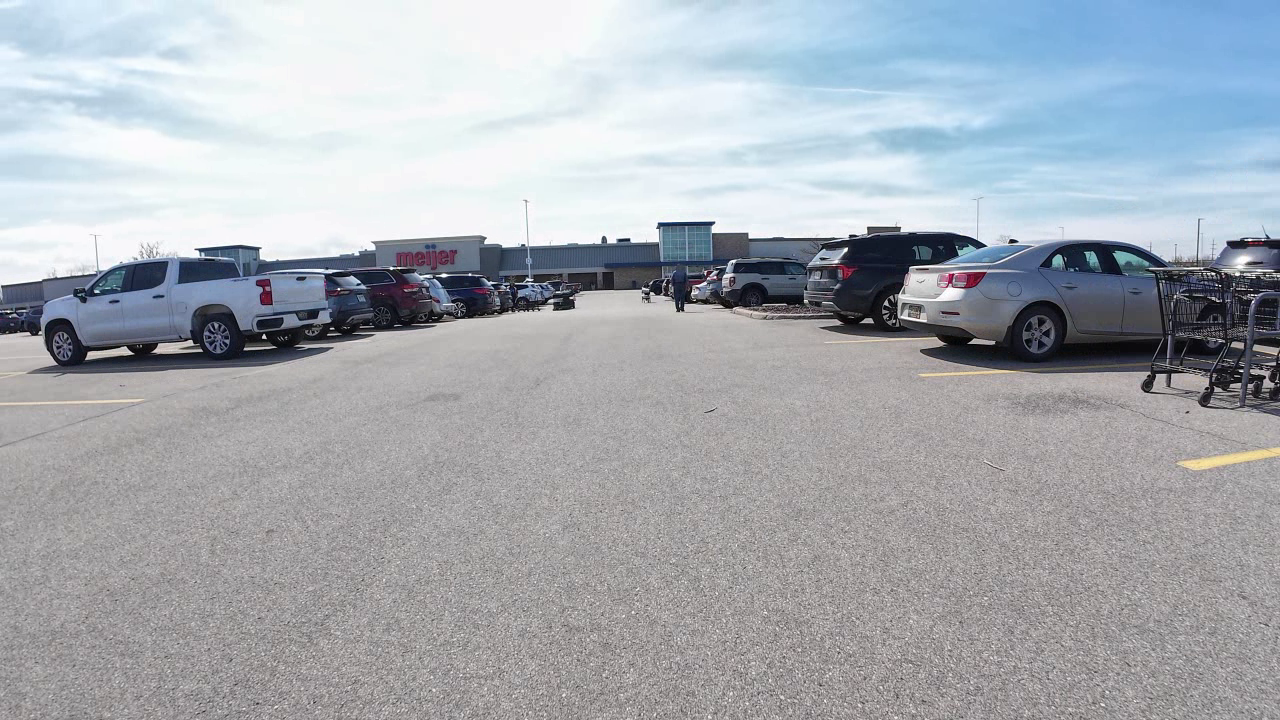}};
\node[label, above=1.5mm of clean] {Clean reference};

\node[img, anchor=center] (fog) at ($(clean.east)+(0.125\textwidth+7mm,0.073\textwidth)$)
{\includegraphics[width=0.245\textwidth]{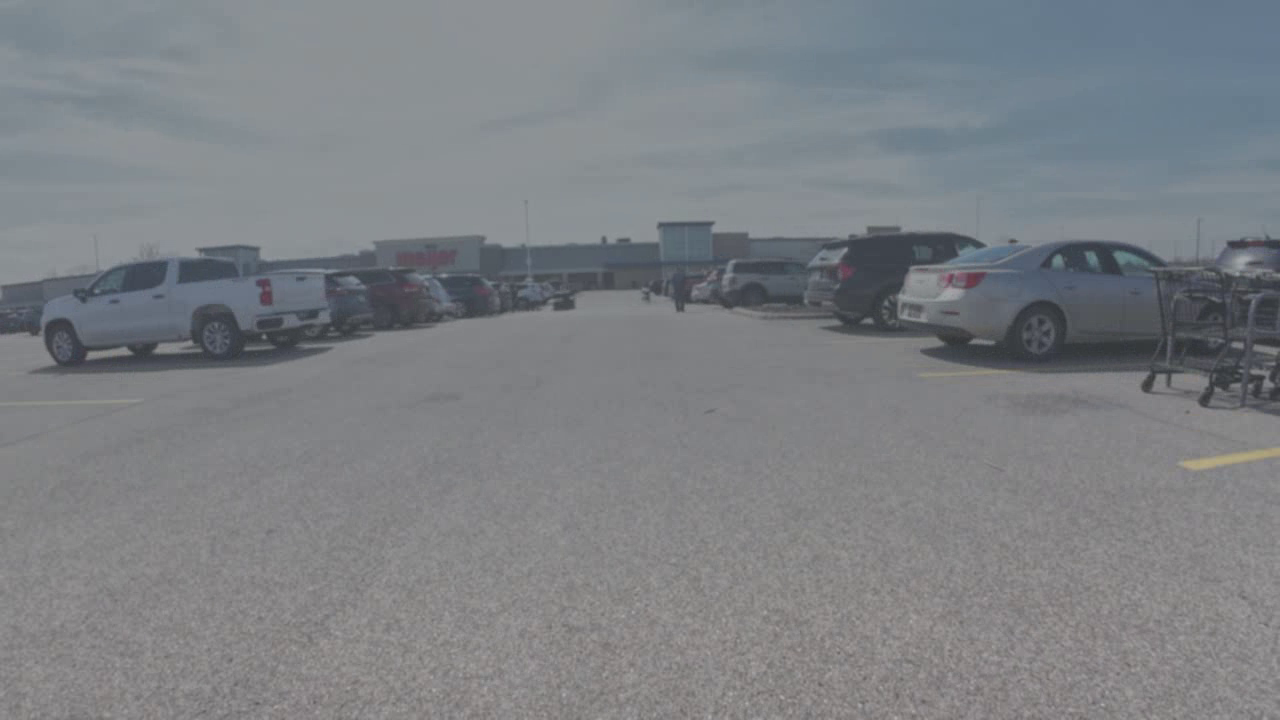}};
\node[label, above=0.8mm of fog] {Fog, severity 5};

\node[img, anchor=center, right=3mm of fog] (rain)
{\includegraphics[width=0.245\textwidth]{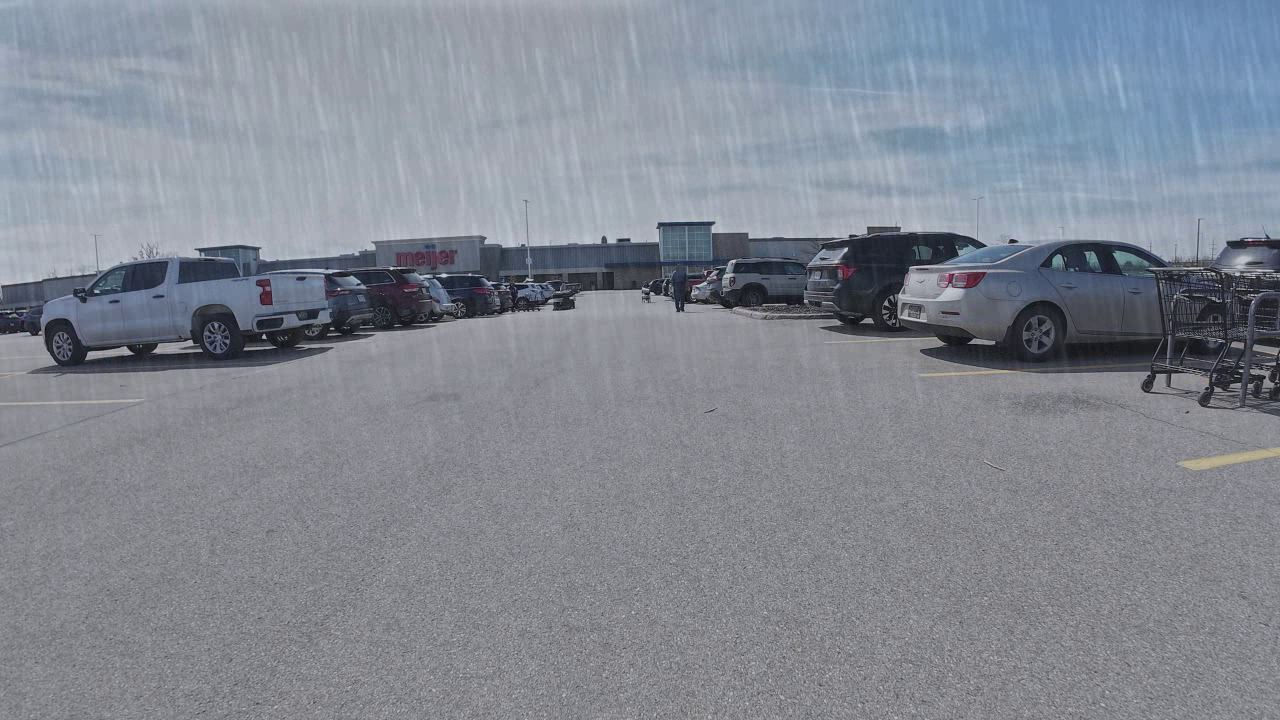}};
\node[label, above=0.8mm of rain] {Rain, severity 5};

\node[img, anchor=center, below=6mm of fog] (motion)
{\includegraphics[width=0.245\textwidth]{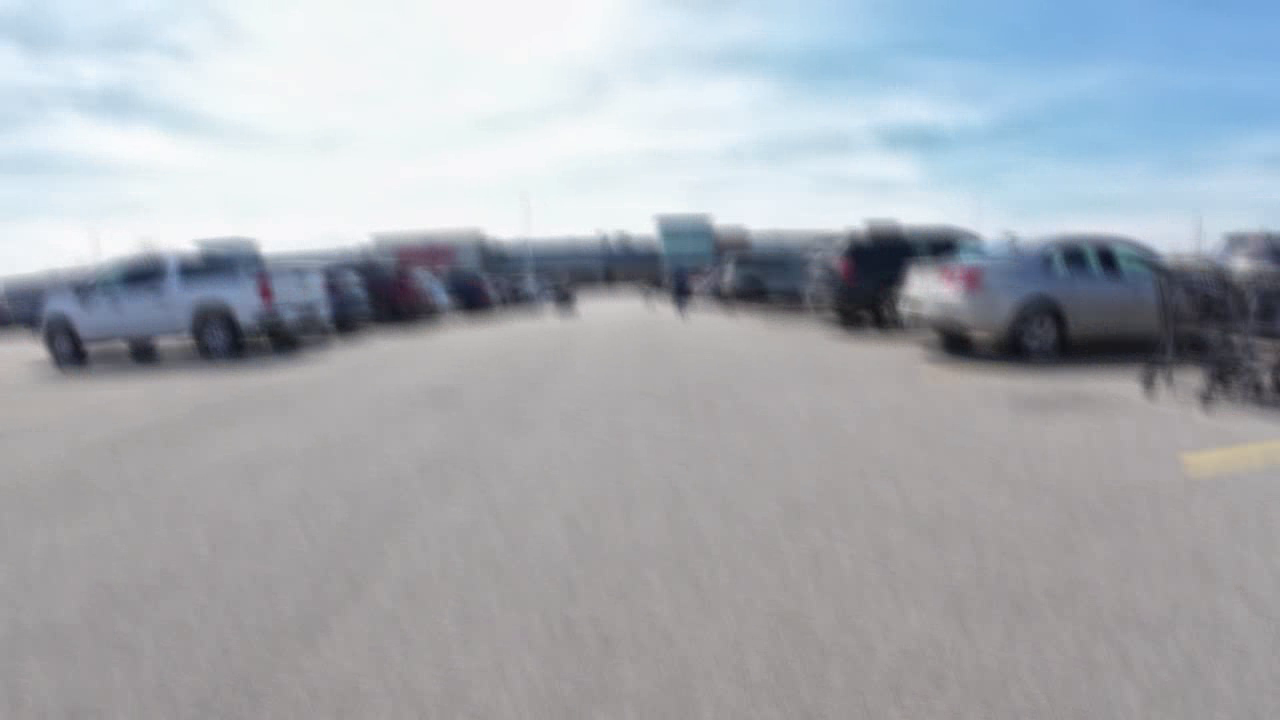}};
\node[label, below=0.8mm of motion] {Motion blur, severity 5};

\node[img, anchor=center, right=3mm of motion] (occ)
{\includegraphics[width=0.245\textwidth]{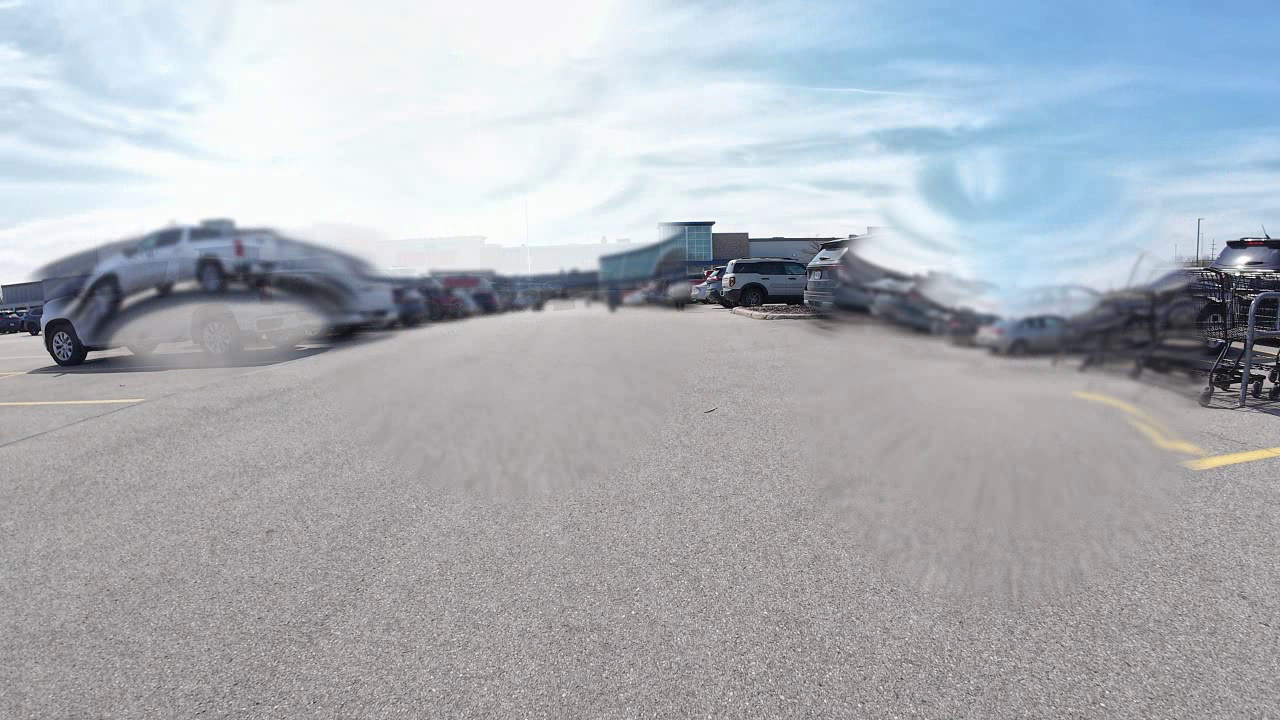}};
\node[label, below=0.8mm of occ] {Lens occlusion, severity 5};

\node[
    draw,
    dashed,
    rounded corners=2mm,
    thick,
    inner sep=5.8mm,
    fit=(fog)(rain)(motion)(occ)
] (corruptbox) {};

\node[
    align=center,
    font=\small\bfseries,
    above=1mm of corruptbox
] {Controlled corrupted variants};

\end{tikzpicture}

\caption{
Teaser of DRIVE-C. A clean forward-facing driving clip is paired with controlled corrupted variants across multiple camera degradation types. DRIVE-C contains 10 scenarios, 12 corruption types, 5 severity levels, 610 clips, 78{,}080 frames, and metadata including severity and Global Sensor Health Index (GSHI) labels.
}
\label{fig:drivec_teaser}
\vspace{-0.5em}
\end{figure*}

\begin{figure*}[t]
\centering
\begin{tikzpicture}[
    font=\small,
    >=Latex,
    box/.style={
        draw,
        rounded corners=2mm,
        thick,
        align=center,
        minimum height=11mm,
        minimum width=26mm,
        fill=blue!5
    },
    dbox/.style={
        draw,
        rounded corners=2mm,
        thick,
        align=center,
        minimum height=11mm,
        minimum width=27mm,
        fill=orange!10
    },
    mbox/.style={
        draw,
        rounded corners=2mm,
        thick,
        align=center,
        minimum height=11mm,
        minimum width=31mm,
        fill=green!8
    },
    outbox/.style={
        draw,
        rounded corners=2mm,
        thick,
        align=center,
        minimum height=12mm,
        minimum width=35mm,
        fill=purple!8
    },
    group/.style={
        draw,
        dashed,
        rounded corners=2mm,
        thick,
        inner sep=4mm
    },
    arrow/.style={
        thick,
        -{Latex[length=2.4mm,width=1.8mm]}
    },
    note/.style={
        align=center,
        font=\footnotesize
    }
]

\node[box] (video) at (0,0) {Real driving\\videos};
\node[box] (anon) at (3.0,0) {Privacy\\anonymization};
\node[box] (clean) at (6.0,0) {Clean clips\\10 scenarios};
\node[dbox] (corr) at (9.2,0) {Controlled\\corruptions};
\node[outbox] (drivec) at (12.7,0) {\textbf{DRIVE-C}\\610 clips\\78,080 frames};

\draw[arrow] (video) -- (anon);
\draw[arrow] (anon) -- (clean);
\draw[arrow] (clean) -- (corr);
\draw[arrow] (corr) -- (drivec);

\node[note] (taxlabel) at (4.3,-3.15) {\textbf{12 degradation modes across 5 severity levels}};

\node[mbox] (weather) at (0.3,-2.0)
{Weather\\fog, rain, \\snow, glare};

\node[mbox] (optical) at (4.8,-2.0)
{Optical\\defocus blur, motion blur, \\low light, occlusion};

\node[mbox] (sensor) at (10.1,-2.0)
{Sensor\\sensor noise, overexposure,\\underexposure, JPEG compression};


\draw[arrow] (corr.south) -- ++(0,-0.35) -| (weather.north);
\draw[arrow] (corr.south) -- ++(0,-0.35) -| (optical.north);
\draw[arrow] (corr.south) -- ++(0,-0.35) -| (sensor.north);

\node[note] (metalabel) at (9.4,-4.6) {\textbf{Released metadata}};

\node[box, fill=yellow!12, minimum width=30mm] (meta1) at (11.0,-3.6) {corruption type\\severity};
\node[box, fill=yellow!12, minimum width=30mm] (meta2) at (11.0,-5.5) {GSHI labels\\clip metadata};


\draw[arrow] ($(drivec.south)!0.38!(drivec.south east)$)  -- ++(0,-3.9) -| (meta1.south);
\draw[arrow] ($(drivec.south)!0.38!(drivec.south east)$) -- ++(0,-3.9) -| (meta2.north);

\node[note] at (7.0,-6.2) {\textbf{Supported research tasks}};

\node[outbox, fill=red!6] (robust) at (2.0,-7.35) {Robustness\\benchmarking};
\node[outbox, fill=red!6] (health) at (5.7,-7.35) {Sensor health\\monitoring};
\node[outbox, fill=red!6] (uncert) at (9.4,-7.35) {Uncertainty\\estimation};
\node[outbox, fill=red!6] (ood) at (13.1,-7.35) {OOD / failure\\analysis};


\draw[arrow] ($(drivec.south)!0.38!(drivec.south east)$) -- ++(0,-5.7) -| (robust.north);
\draw[arrow] ($(drivec.south)!0.38!(drivec.south east)$) -- ++(0,-5.7) -| (health.north);
\draw[arrow] ($(drivec.south)!0.38!(drivec.south east)$) -- ++(0,-5.7) -| (uncert.north);
\draw[arrow] ($(drivec.south)!0.38!(drivec.south east)$) -- ++(0,-5.7) -| (ood.north);

\end{tikzpicture}
\caption{Overview of DRIVE-C. Real forward-facing driving videos are anonymized and converted into clean clips, then transformed using controlled camera corruptions across weather, optical, and sensor/pipeline degradation families. Each corrupted clip includes severity metadata and GSHI labels, enabling robustness benchmarking, uncertainty estimation, and sensor health monitoring for autonomous driving perception.}
\label{fig:drivec_overview}
\end{figure*}
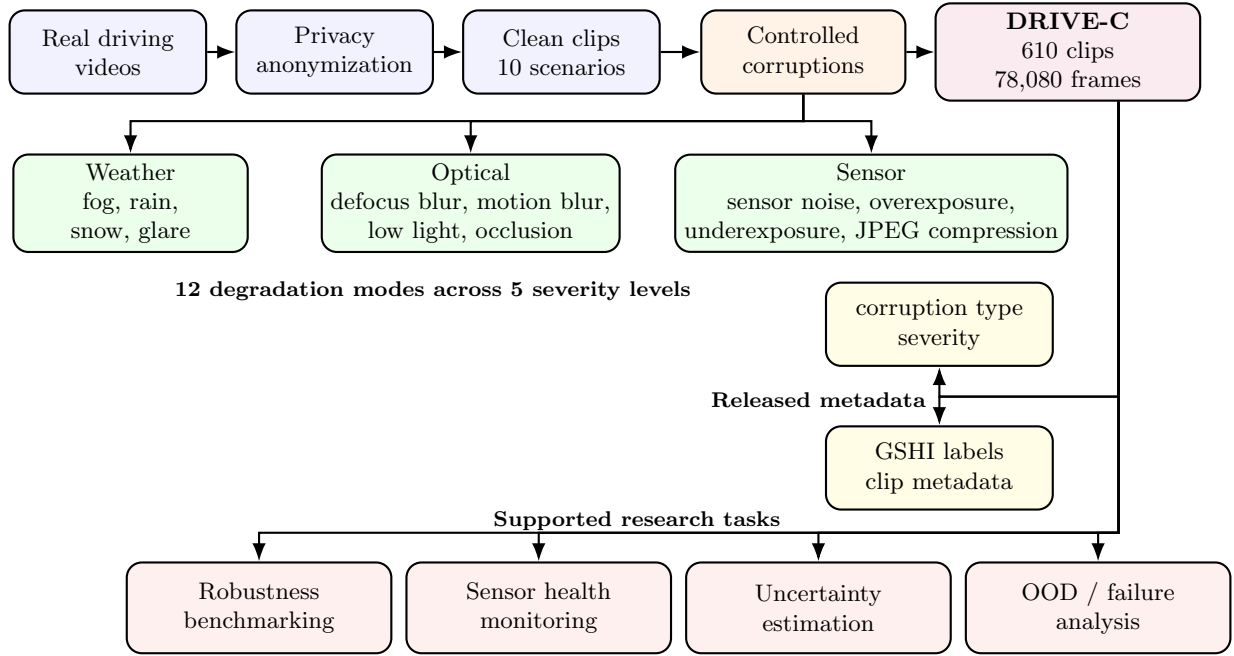

\section*{BACKGROUND}

Camera-based perception is a core component of Advanced Driver Assistance 
Systems (ADAS) and autonomous driving, yet its performance is highly sensitive 
to image degradations encountered in real-world operation. These include 
low-light conditions, glare, motion blur, sensor noise, adverse weather, 
compression artifacts, and partial occlusions. While widely used driving 
datasets such as KITTI~\cite{geiger2013kitti}, nuScenes~\cite{caesar2020nuscenes}, 
and Waymo~\cite{sun2020waymo} support perception tasks under natural conditions, 
they do not provide systematic control over degradation types and severity. 
Conversely, corruption benchmarks such as ImageNet-C~\cite{hendrycks2019imagenetc} 
are typically derived from web imagery and do not preserve the viewpoint, geometry, 
and temporal characteristics of onboard automotive cameras.

DRIVE-C is designed to bridge this gap by combining real-world driving footage with controlled, reproducible degradations. Clean video clips are drawn from forward-facing vehicle-mounted camera recordings across diverse environments, including urban, suburban, freeway, parking, daytime, and nighttime scenarios. A structured set of degradations is then generated using physics-inspired models at multiple severity levels, enabling controlled robustness evaluation while retaining realistic scene characteristics.

The dataset is intended to support a broad range of data-centric research tasks. 
These include robustness benchmarking of perception models under domain shift, 
degradation-aware training and fine-tuning, out-of-distribution (OOD) detection, 
uncertainty calibration, failure mode analysis, and sensor health monitoring for 
safety-critical ADAS applications. The availability of aligned clean and degraded 
clip pairs with structured metadata enables reproducible evaluation and direct 
comparison across degradation conditions and model architectures. Researchers 
developing or evaluating object detectors, lane estimators, depth predictors, or 
semantic segmentation models under realistic sensor degradation scenarios will 
find DRIVE-C directly applicable.

Recent dataset descriptor publications further emphasize the importance of 
structured, well-documented datasets for reproducible research~\cite{makonin2024sfu}. 
Comparable corruption datasets in adjacent domains, such as ImageNet-C, have become 
standard benchmarks within their communities; DRIVE-C aims to serve a similar role 
for automotive camera perception.

This dataset has not previously been used in any publication. The current 
release covers 10 capture scenarios and 12 corruption types, each applied at 5 
severity levels. Detailed dataset statistics are provided in the Records and 
Storage section. The DRIVE-C dataset is publicly available on Zenodo~\cite{aher2026drivec}.

\section*{COLLECTION METHODS AND DESIGN}

\subsection*{Data Acquisition Hardware and Camera Configuration}

DRIVE-C is collected using a forward-facing action camera mounted inside a
passenger vehicle to approximate the viewpoint of a windshield-mounted
automotive perception sensor. The capture platform is a DJI Osmo Action
4~\cite{dji}, installed in a fixed forward-facing configuration to ensure
viewpoint consistency across recording sessions. The camera records natively
at 3840$\times$2160 (4K UHD) and 30\,fps; all clips are downsampled to
1280$\times$720 (HD) for release to reduce storage overhead while retaining
sufficient spatial resolution for perception tasks.

Camera parameters are manually controlled where possible to reduce
inter-session variability. For nighttime driving, representative settings
include a shutter speed of 1/40\,s, ISO 1600, and white balance of 4000\,K.
For daytime driving, lower ISO values and faster shutter speeds are used, with
fixed white balance to avoid frame-to-frame color variation. Cloudy and
transitional lighting conditions are adjusted to reduce highlight clipping and
washed-out sky regions. In-camera low-light enhancement is disabled to preserve
native sensor characteristics.

\subsection*{Capture Scenarios}

Data are collected from real-world driving sessions across diverse environments
in Michigan, United States, on April~1 and April~5, 2026. Environments include
urban streets, suburban roads, rural roads, multi-lane freeways, and a parking
lot. Many recordings contain continuous transitions between scene types. As a
result, DRIVE-C does not rely on rigid per-frame scene labels as primary
metadata fields; instead, each clip is associated with a high-level scene type
and capture condition metadata, enabling flexible downstream labeling.

The dataset comprises 10 scenarios (S01--S10), each a fixed 128-frame clip
(approximately 4.3\,s at 30\,fps) drawn from a distinct source video. The
scenarios span three lighting and weather regimes:

\begin{itemize}
    \item \textbf{Daytime, clear/sunny} --- scenarios S03, S05, S09
          (scene types: urban, parking lot, urban),
    \item \textbf{Daytime, cloudy} --- scenarios S01, S02, S04, S10
          (scene types: urban, freeway, rural, rural),
    \item \textbf{Nighttime, cloudy} --- scenarios S06, S07, S08
          (scene types: urban, urban, urban).
\end{itemize}

Table~\ref{tab:scenarios} summarizes the per-scenario metadata.

\begin{table}[t]
\centering
\caption{Per-scenario metadata for the 10 DRIVE-C capture scenarios.}
\label{tab:scenarios}
\begin{tabular}{@{}llllll@{}}
\toprule
ID  & Scene Type  & Weather & Time    & Traffic  & Frames \\
\midrule
S01 & Urban       & Cloudy  & Day     & Moderate & 128 \\
S02 & Freeway     & Cloudy  & Day     & Moderate & 128 \\
S03 & Urban       & Sunny   & Day     & Moderate & 128 \\
S04 & Rural       & Cloudy  & Day     & Moderate & 128 \\
S05 & Parking lot & Sunny   & Day     & Moderate & 128 \\
S06 & Urban       & Cloudy  & Night   & Moderate & 128 \\
S07 & Urban       & Cloudy  & Night   & Moderate & 128 \\
S08 & Urban       & Cloudy  & Night   & None     & 128 \\
S09 & Urban       & Sunny   & Day     & Low      & 128 \\
S10 & Rural       & Cloudy  & Day     & Moderate & 128 \\
\bottomrule
\end{tabular}
\end{table}

\subsection*{Data Processing Pipeline}

The dataset is constructed using a reproducible multi-stage pipeline:

\begin{enumerate}
    \item Record forward-facing driving video at 4K/30\,fps using a fixed
          vehicle-mounted camera.
    \item Select a 128-frame clip from each source video, downsampled to
          1280$\times$720.
    \item Organize clips by scenario identifier in a structured directory
          hierarchy.
    \item Apply automated anonymization to faces and license plates.
    \item Record processing provenance in per-frame and per-clip metadata files.
    \item Perform quality review to confirm scene coverage and lighting
          representativeness.
    \item Apply 12 controlled corruption types at 5 severity levels using a
          reproducible degradation pipeline to produce the final corrupted
          clip set.
\end{enumerate}

This design separates real-world data acquisition from synthetic corruption
generation and preserves full traceability from each corrupted clip back to
its clean source clip and originating source video.

\subsection*{Anonymization Strategy}

To protect privacy in real-world driving imagery, DRIVE-C employs an automated
anonymization pipeline based on pretrained YOLO-family
detectors~\cite{ultralytics} executed with GPU inference. The anonymization
process includes:
\begin{itemize}
    \item localized elliptical blurring for detected faces,
    \item localized rectangular blurring for detected license plates,
    \item optional fallback anonymization based on broader person or vehicle
          detections when primary detectors yield low confidence.
\end{itemize}

Face blurring is applied using elliptical masks, and license plate blurring is
tightly constrained to detected bounding regions rather than applying coarse
masking, preserving surrounding scene context for downstream perception tasks.

\subsection*{Controlled Corruption Design}

DRIVE-C follows a two-stage design: a clean anonymized base clip set and a
derived corruption clip set. Corruptions are applied exclusively to clean clips
to ensure that privacy processing does not interfere with degradation severity
or introduce confounding artifacts.

The dataset includes 12 corruption types spanning three degradation families,
as listed in Table~\ref{tab:corruptions}. The corruption models are adapted
from the benchmark framework of Hendrycks and
Dietterich~\cite{hendrycks2019imagenetc} and implemented using
OpenCV~\cite{opencv}, with modifications to reflect the geometry, dynamic
range, and temporal characteristics of forward-facing automotive cameras.
Corruption-specific generation parameters (e.g., fog airlight, rain density
and streak angle, motion blur kernel size, JPEG quality factor, and exposure
EV shift) are stored per-clip in the \texttt{extra\_json} metadata field,
enabling full reproducibility of each degraded clip.

Each corruption type is parameterized across five severity levels (1--5) using
a continuous normalized severity value: $s_1 = 0.08$, $s_2 = 0.18$,
$s_3 = 0.35$, $s_4 = 0.55$, $s_5 = 0.75$. Level~1 represents a mild,
barely perceptible degradation; level~5 represents a severe degradation that
substantially impairs scene visibility or sensor readability.

\begin{table}[t]
\centering
\caption{The 12 corruption types in DRIVE-C organized by degradation family.}
\label{tab:corruptions}
\begin{tabular}{@{}ll p{3.3cm}@{}}
\toprule
Family & Corruption Type & Description \\
\midrule
\multirow{4}{*}{Weather}
  & Fog           & Homogeneous scattering reducing contrast and depth cues \\
  & Rain          & Streak-based precipitation with motion-dependent angle \\
  & Snow          & Particle-based accumulation with brightness variation \\
  & Glare / flare & Specular highlight blooming from direct or reflected light \\
\midrule
\multirow{4}{*}{Optical}
  & Motion blur    & Linear kernel blur along the direction of vehicle motion \\
  & Defocus blur   & Radially symmetric blur simulating lens focus error \\
  & Lens occlusion & Partial obstruction from water droplets or debris \\
  & Low light      & Luminance reduction with amplified sensor noise \\
\midrule
\multirow{4}{*}{Sensor}
  & Sensor noise     & Gaussian additive noise applied at the pixel level \\
  & Overexposure     & Global gain increase causing highlight clipping \\
  & Underexposure    & Global gain reduction causing shadow crushing \\
  & JPEG compression & Block-artifact compression at decreasing quality factors \\
\bottomrule
\end{tabular}
\end{table}

\section*{VALIDATION AND QUALITY}

\subsection*{Capture Quality}

The ten clean base clips were reviewed during and after collection to ensure
sufficient visual quality across all capture conditions. Camera settings were
iteratively adjusted per session to reduce common issues such as highlight
clipping, low contrast in cloudy conditions, and excessive motion blur.

Nighttime captures (S06--S08) were validated to confirm that key scene
elements, including lane markings, vehicles, and road structure, remained
visible while preserving realistic sensor noise and low-light characteristics.
Daytime captures were reviewed for sharpness, stable exposure, and
representative coverage of the target environments. The dataset intentionally
preserves natural acquisition variability rather than enforcing strict
normalization, as this variability is relevant to robustness and reliability
evaluation.

The optical and sensor characteristics of the DJI Osmo Action 4 are relevant
to interpreting the clean base clips. The camera uses a 1/1.3-inch CMOS sensor
with a fixed aperture of f/2.8, a 155$^\circ$ ultra-wide field of view, and
supports an ISO range of 100--12800 for video. At the nighttime capture setting
of ISO\,1600, luminance noise is present but controlled, consistent with
manufacturer characterization of the sensor; noise becomes more pronounced above
ISO\,1600, particularly in shadow regions~\cite{dji}. The maximum video bitrate
is 130\,Mbps, and all clips are recorded in H.264/MP4 format prior to the
downsampling step. The camera uses electronic image stabilization (EIS); all
in-camera stabilization and low-light enhancement were disabled during data
collection to preserve native sensor output. These characteristics define the
baseline noise floor and dynamic range of the clean clips from which all
corrupted variants are derived.

\subsection*{Anonymization Quality}

Anonymization quality is recorded for each processed clip. Metadata fields
include per-frame counts of detected face regions, license plate regions, and
fallback anonymization events. This logging enables systematic auditing of privacy
coverage and supports targeted inspection of challenging cases, including small
faces, oblique license plates, nighttime scenes, and motion-blurred frames.

The anonymization pipeline is implemented using OpenCV~\cite{opencv},
PyTorch~\cite{pytorch}, and Ultralytics YOLO models~\cite{ultralytics}.

\subsection*{Near-Duplicate Reduction}

The 10 clean scenarios are selected as fixed 128-frame windows from their
respective source videos, chosen to avoid highly redundant segments such as
stationary stops or prolonged straight-line highway sections with minimal
scene change. This explicit clip-selection approach replaces a general
perceptual-hash deduplication step, providing more deliberate control over
scenario diversity. Perceptual hashing~\cite{imagehash} is available in the
processing codebase for use in future dataset expansions.

\subsection*{GSHI Definition and Ground-Truth Behavior}

The Global Sensor Health Index (GSHI)~\cite{aher2026gshi} is a scalar metric in the range
$[0, 1]$ that quantifies the perceptual impact of a degradation on a given
clip. Ground-truth GSHI values (\texttt{gshi\_gt}) are computed directly
from the applied corruption severity and per-corruption taxonomy weights,
making them fully deterministic given the generation parameters.

\begin{figure}[t]
\centering
\includegraphics[width=\linewidth]{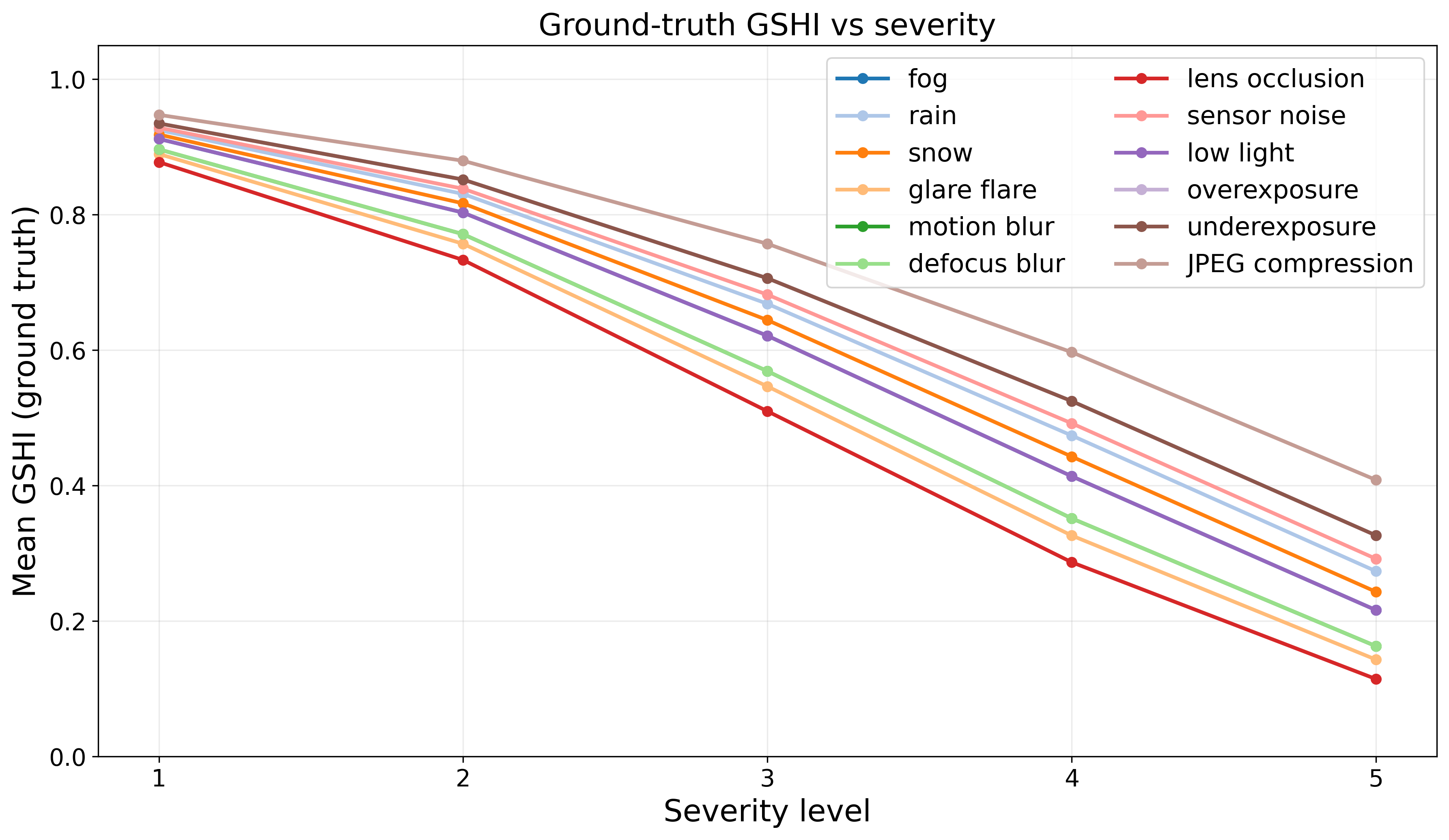}
\caption{Ground-truth GSHI as a function of degradation severity (levels 1--5)
across all 12 corruption types. All corruptions exhibit a consistent monotonic
decrease, confirming the GSHI metric is well-ordered across severity levels.}
\label{fig:gshi_severity}
\end{figure}

As shown in Fig.~\ref{fig:gshi_severity}, \texttt{gshi\_gt} exhibits a
consistent monotonic decrease with increasing severity across all 12
corruption types. This holds across all three degradation families --- weather,
optical, and sensor --- confirming that the GSHI formulation is sensitive to
degradation intensity and well-ordered across severity levels.

\subsection*{Baseline Prediction Analysis}

To assess the practical utility of DRIVE-C as a benchmark, we evaluate
PerceptionHealthNet, a baseline model that directly predicts GSHI from input
clip frames. Across all 610 clips, the Pearson correlation between predicted
GSHI (\texttt{gshi\_pred}) and ground-truth GSHI is $r = 0.339$, with a Spearman
rank correlation of $\rho = 0.341$. The mean predicted GSHI is $0.380$ for clean
clips and $0.283$ for corrupted clips, indicating that the model captures a
coarse separation between clean and corrupted inputs, despite its limited
fine-grained correlation with ground-truth GSHI.

\begin{figure}[t]
\centering
\includegraphics[width=\linewidth]{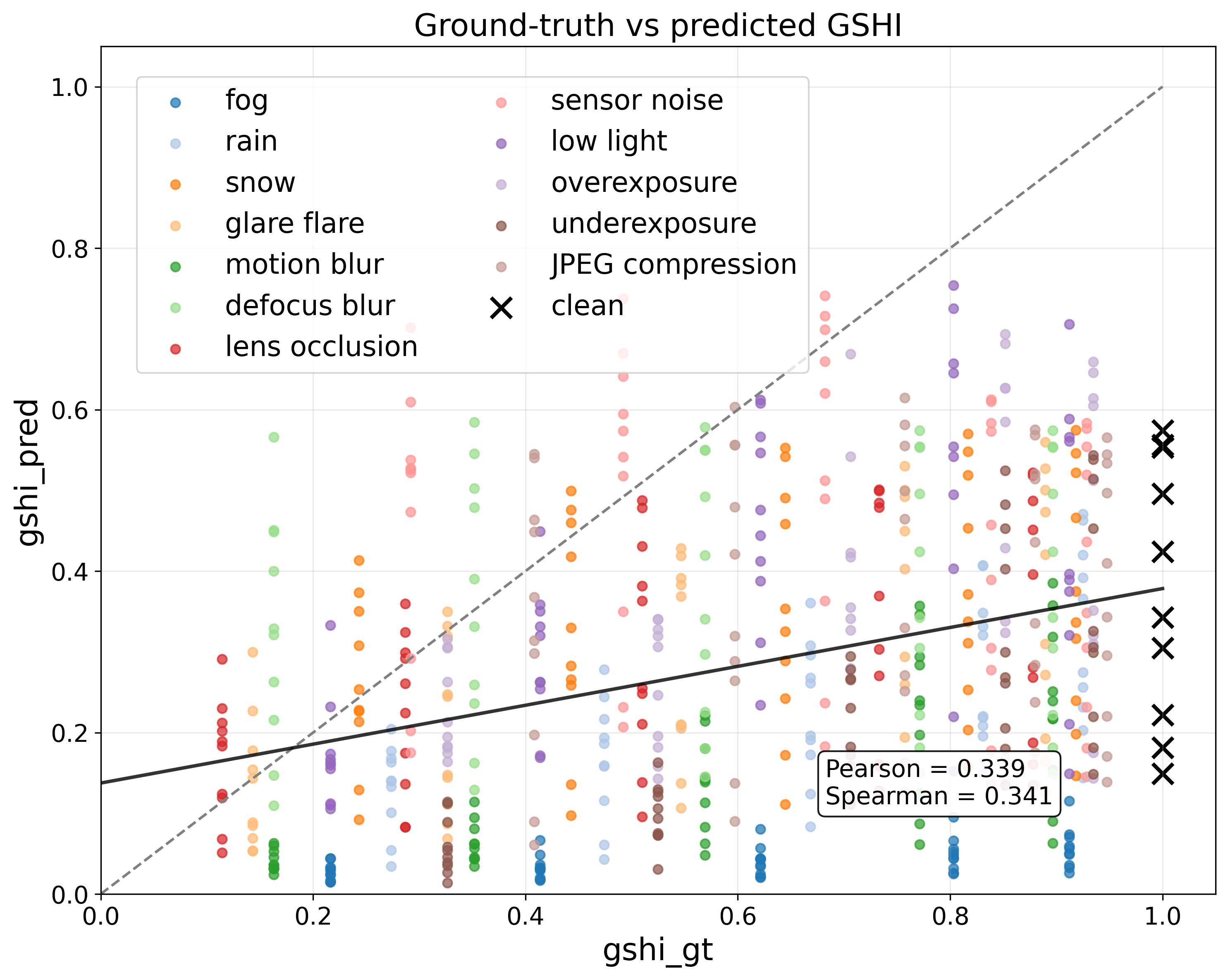}
\caption{Predicted versus ground-truth GSHI across all 610 clips (Pearson
$r = 0.339$, Spearman $\rho = 0.341$). The model shows a systematic bias
toward lower predicted values and increased dispersion at higher ground-truth
GSHI levels.}
\label{fig:gshi_pred_scatter}
\end{figure}

Fig.~\ref{fig:gshi_pred_scatter} shows the scatter of predicted versus
ground-truth GSHI across all samples. Predictions exhibit a systematic bias
toward lower values and increased dispersion, particularly at higher ground-truth
GSHI levels, suggesting the model underestimates perceptual quality for
mild-to-moderate degradations.

\subsection*{Per-Corruption Baseline Results}

Table~\ref{tab:per_corruption} reports per-corruption Pearson and Spearman
correlations, monotonicity of predicted GSHI across severity levels, and mean
predicted GSHI at severities 1, 3, and 5 for all 12 corruption types.

The model performs best on underexposure ($r = 0.77$, $\rho = 0.84$) and
motion blur ($r = 0.73$, $\rho = 0.80$), where predicted GSHI decreases
monotonically and substantially from $s_1$ to $s_5$. Moderate performance is
observed for fog, rain, glare/flare, lens occlusion, low light, and
overexposure ($r = 0.47$--$0.64$). The model fails to track severity for
sensor noise ($r = -0.16$), JPEG compression ($r = 0.09$), defocus blur ($r
= 0.12$), and snow ($r = 0.29$), where predictions remain nearly flat across
severity levels. Only 6 of 12 corruption types show globally monotonic
predicted GSHI from $s_1$ to $s_5$ (monotonicity fraction: 0.475).

These results confirm that DRIVE-C exposes substantial headroom for
degradation-aware modeling, particularly for perceptually subtle corruptions
that the baseline cannot reliably rank.

\begin{table*}[t]
\centering
\caption{Per-corruption baseline evaluation of GSHI prediction across all
12 corruption types (50 clips each). Pearson $r$ and Spearman $\rho$
measure correlation with ground-truth GSHI. Mono.\ indicates whether
predicted GSHI is monotonically non-increasing from $s_1$ to $s_5$.
$\bar{p}_{s1}$, $\bar{p}_{s3}$, $\bar{p}_{s5}$ are mean predicted GSHI
at each severity.}
\label{tab:per_corruption}
\begin{tabular}{@{}lcccrrrr@{}}
\toprule
Corruption Type & Pearson $r$ & Spearman $\rho$ & Mono. &
$\bar{p}_{s1}$ & $\bar{p}_{s3}$ & $\bar{p}_{s5}$ \\
\midrule
Underexposure    & \textbf{0.77} & \textbf{0.84} & Yes & 0.353 & 0.208 & 0.059 \\
Motion blur      & 0.73          & 0.80          & Yes & 0.229 & 0.135 & 0.041 \\
Rain             & 0.64          & 0.65          & Yes & 0.322 & 0.226 & 0.132 \\
Low light        & 0.60          & 0.56          & No  & 0.426 & 0.460 & 0.171 \\
Glare / flare    & 0.56          & 0.55          & Yes & 0.355 & 0.286 & 0.135 \\
Overexposure     & 0.51          & 0.45          & No  & 0.434 & 0.368 & 0.230 \\
Fog              & 0.50          & 0.55          & Yes & 0.057 & 0.041 & 0.028 \\
Lens occlusion   & 0.47          & 0.45          & Yes & 0.341 & 0.309 & 0.167 \\
Snow             & 0.29          & 0.27          & No  & 0.372 & 0.354 & 0.259 \\
Defocus blur     & 0.12          & 0.13          & No  & 0.380 & 0.378 & 0.325 \\
JPEG compression & 0.09          & 0.07          & No  & 0.372 & 0.386 & 0.333 \\
Sensor noise     & $-$0.16       & $-$0.17       & No  & 0.388 & 0.522 & 0.457 \\
\bottomrule
\end{tabular}
\end{table*}

\subsection*{Known Limitations}

DRIVE-C comprises 10 scenarios drawn from video collected on two recording
sessions, which limits diversity in geographic coverage, vehicle type, and
weather conditions. The dataset focuses on forward-facing monocular camera
data and does not include synchronized lidar, radar, or stereo modalities.
Dense semantic annotations such as per-frame bounding boxes or segmentation
masks are not provided. The baseline model (PerceptionHealthNet) shows limited
sensitivity to perceptually subtle corruptions such as sensor noise, JPEG
compression, and defocus blur, which the dataset exposes as open research
challenges. DRIVE-C is therefore most suitable for robustness evaluation,
degradation-aware modeling, and sensor health monitoring research rather than
supervised perception training requiring dense labels. Together, these results
demonstrate that DRIVE-C provides a structured and challenging benchmark for
advancing degradation-aware perception and sensor health estimation in
autonomous driving systems.

\section*{RECORDS AND STORAGE}

DRIVE-C is a video clip dataset. All clips are stored as MP4 files at
1280$\times$720 resolution and 30\,fps. The dataset comprises 610 clips in
total: 10 clean clips and 600 corrupted clips (12 corruption types $\times$
5 severity levels $\times$ 10 scenarios). Each clip is 128 frames
($\approx$4.3\,s). The total frame count across all clips is 78,080
(1,280 clean; 76,800 corrupted). Clips are partitioned into \texttt{dev}
and \texttt{test} splits of 305 clips each, stratified across corruption
types and scenarios.

Table~\ref{tab:dataset_summary} summarizes the top-level dataset statistics.

\begin{table}[t]
\centering
\caption{Top-level DRIVE-C dataset statistics.}
\label{tab:dataset_summary}
\begin{tabular}{@{}ll@{}}
\toprule
Property & Value \\
\midrule
Total clips          & 610 \\
Clean clips          & 10 \\
Corrupted clips      & 600 \\
Scenarios            & 10 (S01--S10) \\
Corruption types     & 12 \\
Severity levels      & 5 (s1--s5) \\
Frames per clip      & 128 \\
Total frames         & 78,080 \\
Resolution           & 1280$\times$720 (HD) \\
Native resolution    & 3840$\times$2160 (4K UHD) \\
Frame rate           & 30\,fps \\
Format               & MP4 (H.264) \\
Dev / test split     & 305 / 305 clips \\
Capture dates        & April 1 and April 5, 2026 \\
Capture location     & Michigan, United States \\
\bottomrule
\end{tabular}
\end{table}

\subsection*{Directory Structure}

The dataset is organized into two top-level directories. Clean clips are
stored flat under \texttt{clean\_clips/}. Corrupted clips are organized
hierarchically by corruption type and then severity level under
\texttt{corrupted/}, making it straightforward to load all clips of a given
corruption type or a given severity across corruption types.

\begin{verbatim}
drive_c/
  clean_clips/
    S01_clean.mp4
    S02_clean.mp4
    ...
    S10_clean.mp4
  corrupted/
    fog/
      s1/  S01_fog_s1.mp4 to S10_xx.mp4
      s2/  S01_fog_s2.mp4 to S10_xx.mp4
      ...
      s5/  S01_fog_s5.mp4 to S10_xx.mp4
    rain/
      s1/ ... s5/
    snow/
      s1/ ... s5/
    [one subdirectory per corruption type]
\end{verbatim}

Each corrupted clip filename follows the convention
{\small\texttt{[scenario\_id]\_[corruption\_type]\_[severity\_name].mp4}}
(e.g., \mbox{\texttt{S03\_motion\_blur\_s4.mp4}}), enabling unambiguous
identification without consulting the metadata file.

\subsection*{Metadata Files}

All metadata is provided in two CSV files distributed with the dataset.

\textbf{\texttt{final\_metadata.csv}} is the primary metadata file with one
row per clip (610 rows, 27 fields). It covers clip identity, corruption
labels, severity values, scene context, video properties, generation
parameters, ground-truth GSHI, and baseline model predictions.
Table~\ref{tab:final_metadata} lists the key fields.

\textbf{\texttt{scenario\_metadata.csv}} provides per-scenario source video
details including the source video identifier, clip frame range, capture
date, original resolution, and scene annotations (10 rows).

\begin{table*}[t]
\centering
\caption{Key fields in \texttt{final\_metadata.csv} (610 rows, 27 fields).}
\label{tab:final_metadata}
\begin{tabular}{@{}lll@{}}
\toprule
Field & Type & Description \\
\midrule
\texttt{sample\_id} & String & Unique clip identifier, e.g., \texttt{S01\_fog\_s3}. \\
\texttt{scenario\_id} & String & Scenario identifier (\texttt{S01}--\texttt{S10}). \\
\texttt{source\_video\_id} & String & Source video identifier for traceability. \\
\texttt{split} & String & Benchmark split: \texttt{dev} or \texttt{test}. \\
\texttt{clip\_type} & String & \texttt{clean} or \texttt{corrupted}. \\
\texttt{corruption\_type} & String & Corruption name, e.g., \texttt{fog}, \texttt{motion\_blur}. \\
\texttt{severity\_level} & Integer & Integer severity (0 = clean, 1--5 = corrupted). \\
\texttt{severity\_name} & String & Severity label, e.g., \texttt{s0}, \texttt{s1}, ..., \texttt{s5}. \\
\texttt{severity\_value} & Float & Continuous severity used by the generator. \\
\texttt{is\_clean} & Boolean & Indicates whether the sample is an uncorrupted clean clip. \\
\texttt{gshi} & Float & Assigned Global Sensor Health Index value for the clip. \\
\texttt{weather} & String & Scene weather, e.g., \texttt{sunny} or \texttt{cloudy}. \\
\texttt{time\_of\_day} & String & Lighting regime: \texttt{day} or \texttt{night}. \\
\texttt{scene\_type} & String & Scene label: \texttt{urban}, \texttt{rural}, \texttt{freeway}, etc. \\
\texttt{traffic\_level} & String & Approximate traffic density, e.g., \texttt{none}, \texttt{low}, or \texttt{moderate}. \\
\texttt{fps} & Integer & Frame rate (30 for all clips). \\
\texttt{resolution} & String & Clip resolution (\texttt{1280x720} for all clips). \\
\texttt{num\_frames} & Integer & Number of frames (128 for all clips). \\
\texttt{output\_path} & String & Relative path to the MP4 file within the dataset. \\
\texttt{extra\_json} & JSON & Per-clip corruption-generation parameters for reproducibility. \\
\texttt{gshi\_gt} & Float & Ground-truth GSHI computed from severity and taxonomy weights. \\
\texttt{gshi\_pred} & Float & Baseline model predicted GSHI. \\
\texttt{pred\_top1\_issue} & String & Highest-probability predicted degradation class. \\
\texttt{pred\_top1\_prob} & Float & Probability of the top predicted class. \\
\texttt{pred\_presence\_json} & JSON & Predicted presence probabilities for all degradation classes. \\
\texttt{pred\_severity\_json} & JSON & Predicted severity values for all degradation classes. \\
\texttt{pred\_present\_thresh\_json} & JSON & Thresholded predicted-present flags for degradation classes. \\
\bottomrule
\end{tabular}
\end{table*}

\subsection*{Clip Relationships}

Each clean clip \texttt{S\textit{XX}\_clean.mp4} is the direct source for
60 corrupted clips (12 corruption types $\times$ 5 severity levels). All
corrupted clips derived from the same scenario are pixel-aligned to their
clean counterpart, enabling paired clean--corrupted evaluation. The
\texttt{scenario\_id} and \texttt{source\_video\_id} fields in
\texttt{final\_metadata.csv} provide full traceability from any corrupted
clip back to its clean base and originating source video.

\balance
\section*{INSIGHTS AND NOTES}

\subsection*{Suggested Usage Patterns}

DRIVE-C supports several distinct research workflows beyond the robustness
benchmarking use case described in the Validation section.

\textbf{Paired clean--corrupted evaluation.} Each corrupted clip has a
pixel-aligned clean counterpart under the same scenario identifier. This
enables direct subtraction-based analysis of degradation effects, no-reference
versus full-reference quality metric comparison, and controlled ablation
of model components sensitive to specific corruption families.

\textbf{Severity-aware training and fine-tuning.} The continuous
\texttt{severity\_value} field ($s_1 = 0.08$ through $s_5 = 0.75$) can
be used as a regression target or curriculum signal for training
degradation-aware models, beyond the five discrete severity levels.

\textbf{Multi-label degradation detection.} The \texttt{pred\_presence\_json}
and \texttt{pred\_severity\_json} fields provide per-clip probability
estimates across all 12 degradation classes, supporting multi-label
classification and degradation mixture studies.

\textbf{Time-series and temporal analysis.} Each clip is a contiguous
128-frame sequence at 30\,fps. Researchers studying temporal dynamics of
degradation effects, optical flow robustness, or video quality assessment
can exploit the temporal structure rather than treating clips as independent
frame collections.

\textbf{General video quality assessment.} Although DRIVE-C is designed for
automotive perception, the controlled degradation structure and GSHI metric
are not specific to driving. The dataset can serve as a benchmark for
no-reference video quality estimators in any domain where camera sensor
health is relevant.

\subsection*{Caveats and Special Cases}

\textbf{Scene type imbalance.} The 10 scenarios are not uniformly distributed
across scene types: 6 are urban, 2 rural, 1 freeway, and 1 parking lot.
Urban scenarios are overrepresented, and scene types such as tunnels,
construction zones, and highway on-ramps are absent. Researchers drawing
conclusions about freeway or parking-lot-specific robustness should account
for this imbalance.

\textbf{Nighttime scenario concentration.} All three nighttime scenarios
(S06, S07, S08) are urban and cloudy. There are no nighttime rural, freeway,
or clear-sky scenarios in the current release. Models evaluated on nighttime
clips in DRIVE-C are therefore evaluated specifically on urban nighttime
conditions, not nighttime driving in general.

\textbf{Baseline model false positives on clean clips.} The baseline
PerceptionHealthNet model assigns low predicted GSHI to several clean clips,
with nighttime scenarios particularly affected: S08 ($\hat{g} = 0.149$,
predicted class: vignetting) and S07 ($\hat{g} = 0.181$, predicted class:
vignetting) receive lower health scores than many corrupted clips. This
reflects the model's difficulty distinguishing natural low-light acquisition
characteristics from synthetic degradation. Users relying on
\texttt{gshi\_pred} for quality filtering should validate thresholds against
the clean clip distribution before applying them to new data.

\textbf{Non-monotonic baseline predictions for subtle corruptions.}
For sensor noise, JPEG compression, defocus blur, and snow, the baseline
model's predicted GSHI does not decrease monotonically with increasing
severity --- and for sensor noise the Pearson correlation with ground-truth
GSHI is negative ($r = -0.16$). These corruption types should be treated as
open benchmarking challenges rather than solved cases in any downstream
evaluation.

\textbf{Dev/test split assignment.} Scenarios S01--S05 are assigned to the
\texttt{dev} split and S06--S10 to the \texttt{test} split. Because split
assignment is at the scenario level (not the clip level), the dev and test
sets differ not only in sample identity but also in scene type and lighting
distribution: the dev set contains the freeway and parking-lot scenarios,
while the test set contains all three nighttime scenarios. Researchers
should be aware of this structural difference when interpreting generalization
results between splits.

\textbf{Privacy processing and native sensor characteristics.} Corrupted
clips are generated from anonymized clean clips. The anonymization step
(localized face and license plate blurring) modifies small regions of each
frame. For the vast majority of pixels and for global image statistics,
this has no effect. However, studies specifically examining face or vehicle
detection robustness under degradation should note that face and plate
regions have been replaced with blurred content.

\textbf{Corruption independence assumption.} Each corrupted clip contains
exactly one active corruption type. Real-world sensor degradation often
involves simultaneous or correlated degradations (e.g., rain combined with
lens occlusion, or low light combined with motion blur). DRIVE-C's
single-corruption design enables controlled evaluation but does not directly
cover compound degradation scenarios. Researchers interested in compound
degradations can construct them by applying the provided generation scripts
to the clean clips with multiple active corruption types.

\section*{SOURCE CODE AND SCRIPTS}

The DRIVE-C dataset is constructed using a Python-based processing pipeline.
The associated repository provides utilities for:
\begin{itemize}
    \item selecting and extracting scenario clip windows from source videos,
    \item anonymizing faces and license plates using GPU-accelerated 
          YOLO-family detectors,
    \item generating controlled corruption variants across all 12 corruption 
          types and 5 severity levels,
    \item computing ground-truth GSHI values from applied severity and 
          taxonomy weights,
    \item running the PerceptionHealthNet baseline model to produce 
          \texttt{gshi\_pred} and per-class degradation predictions, and
    \item generating and validating the \texttt{final\_metadata.csv} and 
          \texttt{scenario\_metadata.csv} manifest files.
\end{itemize}

The codebase is publicly available at:
\begin{center}
\url{https://github.com/shiv-aher/drive-c-dataset}
\end{center}

The repository includes all scripts and configuration files required to
reproduce the dataset generation process from the original source videos.
Corruption-specific generation parameters for each clip are stored in the
\texttt{extra\_json} field of \texttt{final\_metadata.csv}, enabling any
individual corrupted clip to be regenerated independently without rerunning
the full pipeline.

To support long-term reproducibility, the source code repository will be
archived with a persistent DOI via Zenodo corresponding to the dataset release
version. The dataset itself will be publicly released through a permanent data
repository with open access.

The implementation is written in Python and leverages widely used open-source
libraries including OpenCV~\cite{opencv}, PyTorch~\cite{pytorch}, and
Ultralytics YOLO~\cite{ultralytics}.

\section*{ACKNOWLEDGEMENTS}

The dataset was collected, processed, and documented by the author. The data collection system was designed, the processing pipeline implemented, corruption variants generated, and the dataset prepared for publication independently.

Special thanks to Manisha A.\ for her support in real-world data collection, including coordinating capture sessions, assisting with camera setup, and ensuring consistent coverage across diverse driving environments.

This work was conducted independently, without external grants or awards.

The author declares no conflicts of interest.

\bibliographystyle{IEEEtran}

\bibliography{references}

\end{document}